\documentclass[11pt,a4paper]{article}
\usepackage[hyperref]{acl2020}
\usepackage{times}
\usepackage{latexsym}

\usepackage{amssymb}
\usepackage[ruled,vlined]{algorithm2e}
\usepackage{algorithmicx}

\usepackage{url}
\usepackage{tikz}
\usepackage{pgfplots}
\usepackage{textcomp}
\usepackage{varioref}
\usepackage{graphicx}
\pgfplotsset{width=9cm,compat=1.9}
\usepackage{float}
\usepackage{multirow}
\usepackage{array,booktabs,ragged2e}
\usepackage{amsmath}
\usepackage[bottom]{footmisc}
\usepackage{scalefnt}

\usepackage{booktabs}
\usepackage{pifont}
\usepackage{microtype}

\aclfinalcopy 

\newcolumntype{R}[1]{>{\RaggedLeft\arraybackslash}p{#1}}
\title{Self-Training for Unsupervised Parsing
with PRPN}

\author{
 Anhad Mohananey$^1$\thanks{\ \ Equal contribution.}~\hspace{0.15em}\thanks{\ \ Now at Electronic Arts.}\hspace{0.4em}
 Katharina Kann$^2$\footnotemark[1]\hspace{0.4em}
 Samuel R. Bowman$^1$ \\
\textnormal{$^1$New York University} \\
\textnormal{$^2$University of Colorado Boulder} \\
 \texttt{\{anhad,bowman\}@nyu.edu} \\ 
  \texttt{katharina.kann@colorado.edu}
 \\
{}
 }

\date{}

\begin{document}
\maketitle

\begin{abstract}
Neural unsupervised parsing (UP) models learn to parse without access to syntactic annotations, while being optimized for another task like language modeling. In this work, we propose \emph{self-training} for neural UP models: we leverage aggregated annotations predicted by copies of our model as supervision for future copies. 
To be able to use our model's predictions during training, we extend a recent neural UP architecture, 
the PRPN \citep{shen2017neural}, 
such that it
can be trained in a semi-supervised fashion. We then add examples with parses predicted by our model to our unlabeled UP training data.
Our self-trained model outperforms the PRPN by $8.1\%$ F1 and the previous state of the art by $1.6\%$ F1.
In addition, we show that our architecture can also be helpful for semi-supervised parsing in ultra-low-resource settings.
\end{abstract}

\section{Introduction}

Unsupervised parsing (UP) models learn to parse sentences into unlabeled constituency trees without the need for annotated treebanks. Self-training \citep{Yar95,Riloff+Wiebe+Wilson:03a} consists of training a model, using it to label new examples and, based on a confidence metric, adding a subset to the training set, before repeating training.
For supervised parsing, results with self-training have been mixed \citep{charniak1997statistical,steedman2003bootstrapping,effselftrain}. For unsupervised dependency parsing, \citet{le2015unsupervised} obtain strong results by training a supervised parser on outputs of unsupervised parsing. 
UP models show low self-agreement between training runs \cite{kim2019compound}, while obtaining parsing performances far above chance. Supervising one run with confident parses from the last could combine their individual strengths.
Thus, we ask the question: \textit{Can UP benefit from self-training?}

In order to answer this question, we propose SS-PRPN, a semi-supervised extension of the UP architecture PRPN \citep{shen2017neural}, which can be trained jointly on language modeling and supervised parsing. This enables our model to leverage silver-standard annotations obtained via self-training for supervision. Our approach draws on the idea of \textit{syntactic distances}, which can be learned both as latent variables \citep{shen2017neural} and as explicit supervision targets \citep{shen2018straight}.
We use both of these, leveraging annotations obtained via UP to supervise the two different outputs of the parser, in addition to standard UP training. 

\begin{figure}[t]
\centering
\includegraphics[width=.75\columnwidth]{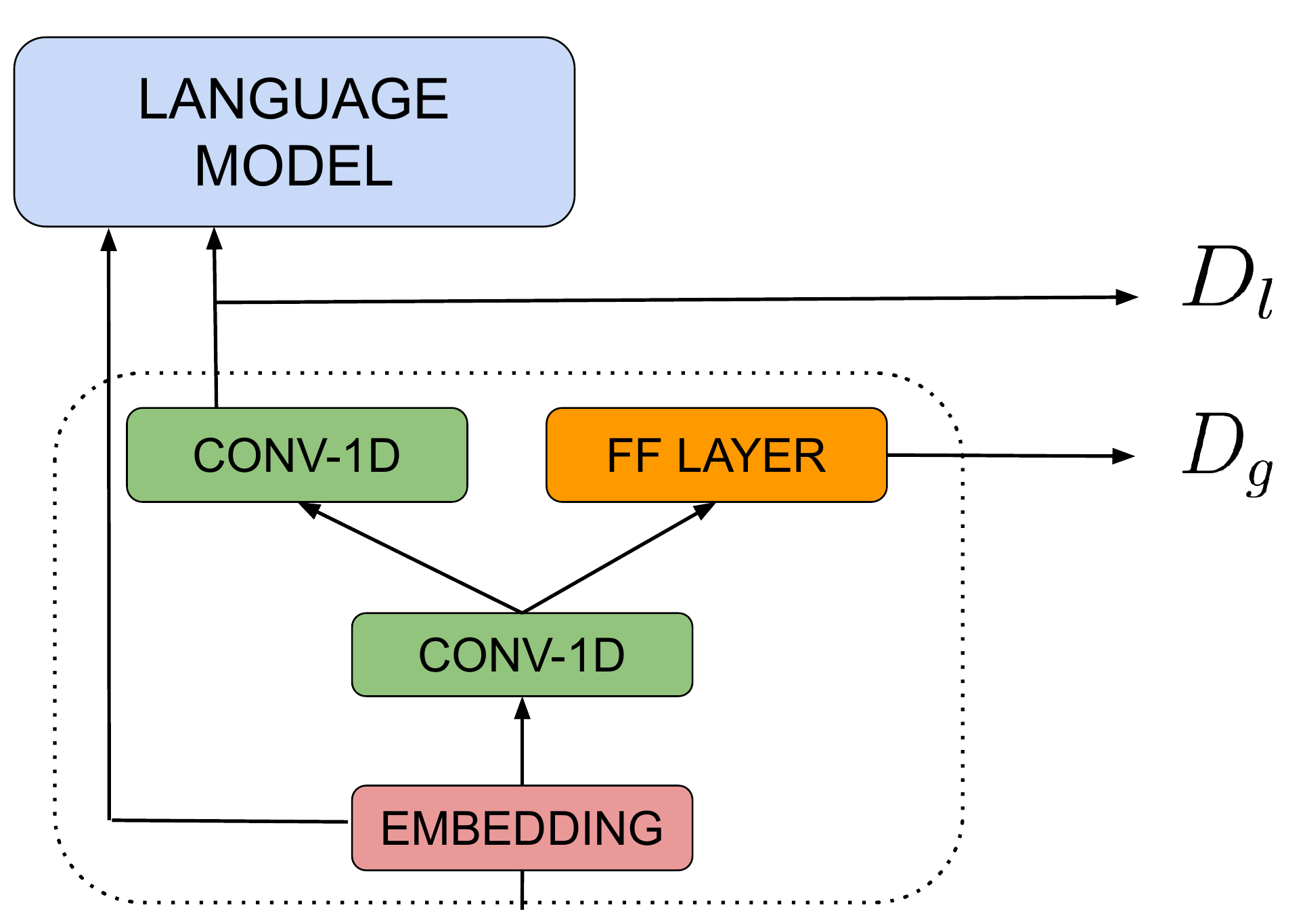}
\caption{Our parser, represented by the dotted box, outputs syntactic distances $D_g$ and $D_l$. Both $D_g$ and $D_l$ can be supervised, but $D_l$ can also be learned in a latent manner.  
}\label{diagram}
\end{figure}

SS-PRPN, in combination with self-training, improves over its original version by $8.1\%$ F1 and over the previous state of the art \citep{kim2019compound} by $1.6\%$ F1, when trained and evaluated on the English PTB \citep{mitchell1999treebank}: \textit{UP can indeed benefit from self-training.} We further perform an analysis of our self-training procedure, finding that longer sentences benefit most from self-training.

Although our primary motivation for the development of a semi-supervised architecture is to enable self-training, we further hypothesize that, since language modeling and parsing annotations seem to provide complementary information, UP should aid low-resource supervised parsing. As a proof of concept, we employ SS-PRPN for semi-supervised training. In extremely-low-data regimes with no more than 250 labeled parses, SS-PRPN outperforms supervised and unsupervised baselines in most settings on unlabeled parsing, and in all settings on labeled constituency parsing. 

\paragraph{Related Work}
Following the line of research on non-neural UP models \citep{clark2001unsupervised,klein2002generative,bod-2006-subtrees}, early approaches to neural UP \citep{yogatama2016learning,choi2018learning} obtain improved performance on downstream tasks, yet show highly inconsistent behavior in parsing \citep{williams2018latent}. 

Recently, \citet{shen2017neural} introduce the first high performing neural UP model \citep{htut2018grammar}. \citet{dyer2019critical} raise concerns that PRPN's parsing methodology is biased towards English trees.
Though these concerns are serious, they are largely orthogonal to our research question regarding the helpfulness of self-training for UP.

Several models have been introduced since: \citet{shen2018ordered} propose 
an architecture consisting of an LSTM \citep{hochreiter1997long} with a modified update function for the LSTM cell state, \citet{kim2019compound}---the current state-of-the-art---introduce a model based on a mixture of probabilistic context-free grammars, \citet{kim2019unsupervised} present unsupervised learning of recurrent neural networks grammars, \citet{li2019imitation} combine PRPN with imitation learning, and \citet{drozdov2019unsupervised} employ a recursive autoencoder. \citet{kim2020pre} examine tree induction from pre-trained models.

\section{Model}

\paragraph{Syntactic Distances} \label{ssec:dist}
In order to parse a sentence, a computational model needs to output some kind of variables representing a unique tree structure.
The variables we use are \textit{syntactic distances} as introduced by \citet{shen2017neural}.
They represent the syntactic relationships between all successive pairs of words in a sentence.
If the distance between two neighboring words is large,
they belong to different subtrees, and, thus, their traversal distance in the tree is large.
A parse tree can be created by finding the maximum syntactic distance, splitting the sentence into sub-trees there, and repeating this process recursively for each sub-tree until a single token is left. 

Two different formulations of syntactic distance have been proposed to realize this basic intuition:
The first, which we refer to as $D_l$, is introduced by \citet{shen2017neural} as a latent variable in their UP model. 
Since, for self-training, we supervise $D_l$ with values predicted by our model, we introduce Algorithm \ref{algol_t_d}, which is used to convert a tree to distances $D_l$.
The second kind of distance, denoted here as $D_g$, is introduced by \citet{shen2018straight} as  labels for direct supervision. We use their algorithms to map trees to distances $D_g$ and vice versa, and ask readers to refer to \citet{shen2018straight} for details.

We design our parser in such a way that it can predict both. 
We treat the decision whether $D_g$ or $D_l$ are used at test time as a hyperparameter. 
The reasons why we employ both types of distances are two-fold: $D_g$, unlike $D_l$, cannot be learned in an unsupervised fashion, which is critical for a semi-supervised architecture. Empirically, supervising purely on $D_l$ performs poorly. 

\LinesNumbered
\begin{algorithm}[t!]
\small
\SetAlgoNoLine
D$_l$ $\leftarrow$ [1] * leaves$_{tree}$ \Comment{leaves$_{tree}$ :leaf count of tree}\\
$b \leftarrow 0$\\
$max \leftarrow 100$\Comment{max: max possible depth of tree}\\
\textbf{Function}
\SetKwFunction{ttod}{$\textrm{DISTANCE}$}{}{}
\ttod{\textrm{tree}, b, max}{\\
      \Indp\textbf{}{
        \ttod{tree$_l$, b, max-1}\\
        x $\leftarrow$ tree$_r$\Comment{tree$_r$: right child of tree}\\
        \While{True}
        {
            \If{x$_l$ is empty}{  
                D$_l$$[$b + leaves$_{tree_{l}}]$ $\leftarrow$ max\\
                \textbf{break}
            }
            x $\leftarrow$ x$_l$\Comment{x$_l$: left child of x}
        }
      }     
      }

 \caption{Tree to latent distances $D_l$}\label{algol_t_d}
 \end{algorithm}

\paragraph{The Parser} \label{parser} 
Our parser, cf. Figure  \ref{diagram}, consists of an embedding layer and a convolutional layer which are followed by two different components: a linear output layer that predicts  $D_g$ and a second convolutional layer that predicts  $D_l$.

Formally, given an input sentence $s = t_0, t_1, \dots, t_{n-1}$, our parser predicts $D_g$ as:
\begin{align}
{h_i} = {ReLU}({W_c
\begin{bmatrix}
    t_{i-L_1}  \\
    t_{i-L_1+1}  \\
    \cdots \\
    t_{i}    
\end{bmatrix}}{ + }b_c) 
\end{align}
\begin{align}
{d_i} = {ReLU}({W_d}{h_i}{ + }b_d)
\end{align}
where $W_c$ are the weights of the first convolutional layer, $W_d$ are the weights of the output layer corresponding to $D_g$, and $b_c$ and $b_d$ are bias vectors. $L_1$ is the filter size.
$D_l$ involves similar computations, but is the output of the second layer.
\paragraph{Distance Loss} \label{loss_dist}
When we have silver-standard annotations from self-training available, we compute the loss for both syntactic distances 
directly. 
Since the relative ranking between distances---rather than absolute values---defines the  tree structure, we train our parser 
with a hinge ranking loss following \citet{shen2018straight}. 
Our distance loss $L_r$ is the weighted sum of the distance losses corresponding to $D_l$ ($L_{sl}$) and $D_g$ ($L_{sg}$): 
\begin{equation}
{L_r} = \alpha  {L_{sg}} + (1 - \alpha)  {L_{sl}}
\end{equation}
\paragraph{Language Modeling Loss} \label{loss_lm}
In order to optimize the parameters of our parser without direct supervision, we further feed its output---the predictions for $D_l$---into a language model, following \citet{shen2017neural}.

\paragraph{Multi-Task Training}
Our parser is trained in a semi-supervised fashion with losses corresponding to (i) learning the distances in a latent manner through \textit{language modeling}, and (ii) supervising directly on \textit{distances}.
We sample batches from both objectives at random.


\begin{algorithm}[t!]
\small
\SetAlgoNoLine
Unlabeled data $X_U$ \\
Training set $X_T \leftarrow \emptyset$ \\
Train $n_c$ UP models on $X_U$\\
\For{$s_i \in X_U$}
 {
     $n_a \leftarrow$ number of models agreeing on parse $p(s_i)$ \\
     \If {$n_a \geq \mu  n_c$}
         {$X_T \leftarrow X_T \cup p(s_i)$ \Comment{add confident parse}} 
 }
 Train model on $X_U$ and $X_T$
 \caption{Self-training for UP}\label{algo2}
 \end{algorithm}
\paragraph{Self-Training}
For self-training, cf. Algorithm \ref{algo2},
we first train $n_c$ models on the unlabeled PTB training set $X_U$.
We then have them predict parse trees for all sentences in $X_U$. If more than $\mu * n_c$ models (with $\mu$ as a hyperparameters) agree on the same parse, we add it as a silver-standard \textit{labeled} example to 
the parsing training set $X_T$.
We use Algorithm \ref{algol_t_d} and the respective algorithm by \citet{shen2018straight} to convert consensus trees into distances $D_l$ and $D_g$. 
We then train a new model on both $X_U$ and $X_T$. 


\section{Experimental Design}
\paragraph{Data and Metrics}
We experiment on the English Penn Treebank \citet[PTB;][]{mitchell1999treebank}.
For evaluation, we compute the F1 score of the output parses against binarized gold parses following \citet{williams2018latent}. The code for our model is published online\footnote{https://github.com/anhad13/SelfTrainingAndLRP}.

\paragraph{Baselines}
We compare against an unsupervised recurrent neural network grammar \citep[URNNG;][]{kim2019unsupervised}, a compound probabilistic context free grammar \citep[C-PCFG;][]{kim2019compound}, and \citet{shen2017neural}'s PRPN. We re-implement and tune PRPN in our code base.







\paragraph{Hyperparameters} \label{ssec:hyper}
We tune our hyperparameters on the development set. 
Hidden states and word embeddings have 300 and 100 dimensions, respectively. We set the weight $\alpha = 0.5$.
For self-training, we obtain best results with $\mu = 60\%$ and $n_c = 15$. We further experiment with converting either $D_l$ or $D_g$ into final parse trees, and find that $D_l$ works best. 

\section{Results and Analysis}

\paragraph{Unsupervised Parsing Performance}
Table \ref{fig:self} shows our results. 
SS-PRPN outperforms all baselines: 
our model 
obtains a $1.6\%$ higher F1 score than the strongest baseline. It further improves substantially over comparable non-self-trained baselines: by 14.6\% over PRPN and by 8.1\% over our reimplementation of it. SS-PRPN also shows a much lower variance. 
This demonstrates that self-training is indeed a viable approach for UP.

\begin{table}[t]
\small
\centering
 \begin{tabular}{l r r}
\toprule
 Model & F1($\mu$)\\ [0.5ex] 
\midrule
 PRPN & 39.8 (5.6)\\
 PRPN (ours) & 46.3 (6.3)\\ 
 C-PCFG & 52.8 (3.8)\\ 
 URNNG  & 44.8 (4.1)\\
 \midrule
 SS-PRPN & \textbf{54.4 (0.6)}\\
 \midrule
 Left Branching (LB) & 13.1\\
 Right Branching (RB)& 16.5\\
 Random & 21.4\\
 \bottomrule
\end{tabular}
\caption{Results on the English PTB test set, with the model tuned on the dev set. LB, RB and Random baselines are taken as-is from \citet{htut2018grammar}. Since evaluation of C-PCFG, PRPN and URNNG is done against binary gold trees, results might differ from the original papers. 
}
\label{fig:self}
\end{table}

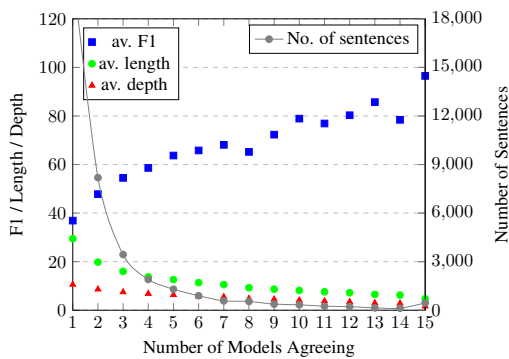
\begin{figure}[]
\centering
\resizebox{0.9\columnwidth}{!}{%
\centering
\resizebox{6.5cm}{!}{
\begin{tikzpicture}
\begin{axis}[
    xlabel={Number of Models Agreeing},
    ylabel={F1 / Length / Depth},
    xmin=1, xmax=15,
    ymin=0, ymax=120,
    xtick={1,2,3,4,5,6,7,8,9,10,11,12,13,14,15},
    ytick={0,20,40,60,80,100, 120},
    legend pos=north west,
    ymajorgrids=true,
    grid style=dashed,
]

\addplot[
    color=blue,
    mark=square*,
    only marks
    ]
    coordinates {
  (1, 36.9)(2, 47.8)(3, 54.5)(4, 58.6)(5, 63.7)(6, 65.8)(7, 68.1)(8, 65.2)(9, 72.3)(10, 78.9)(11, 76.9)(12, 80.3)(13, 85.7)(14, 78.4)(15, 96.5)
    };
    \addlegendentry{av. F1}
\addplot[
    color=green,
    mark=*,
    only marks
    ]
    coordinates {
  (1,29.45)(2, 19.8)(3, 16.0)(4, 13.72)(5, 12.62)(6, 11.37)(7, 10.57)(8, 9.26)(9, 8.67)(10, 8.15)(11, 7.61)(12, 7.21)(13, 6.5)(14, 6.23)(15, 4.73)
    };
    \addlegendentry{av. length}
\addplot[
    color=red,
    mark=triangle*,
    only marks
    ]
    coordinates {
 (1,10.58) (2, 8.61)(3, 7.48)(4, 6.81)(5, 6.35)(6, 5.74)(7, 5.47)(8, 4.76)(9, 4.5)(10, 4.11)(11, 3.77)(12, 3.45)(13, 3.0)(14, 2.83)(15, 1.63)
    };
    \addlegendentry{av. depth}

\end{axis}

\begin{axis}[
  axis y line*=right,
  axis x line=none,
  xmin=1, xmax=15,
  ymin=0, ymax=18000,
  ytick={0,3000,6000,9000,12000,15000,18000},
  scaled y ticks=false,
  ylabel=Number of Sentences,
]

\addplot[smooth,mark=*,gray]
  coordinates{
    (1,20964)
    (2,8191)
    (3,3437)
    (4,1899)
    (5,1298)
    (6,896)
    (7,578)
    (8,541)
    (9,377)
    (10,336)
    (11,247)
    (12,221)
    (13,143)
    (14,121)
    (15,452)

}; \addlegendentry{No. of sentences}
\end{axis}

\end{tikzpicture}
}
}

    


\caption{Statistics for self-training ($n_c=15$): As agreement among UP models goes up, parsing F1 improves, and average depth and length go down.    
}
\label{fig:analysis:best}
\end{figure}

\paragraph{Analysis of Self-Training}
We interpret agreement rate as our confidence value for self-training, with the hypothesis that, as agreement among models increases, there is a higher likelihood that the parse is correct.  
In Figure \ref{fig:analysis:best}, we show that, as expected, the F1 score increases as more models agree, for the best self-training run (15 individual models, or the second last row in Table \ref{fig:self}). 

Additionally, Figure \ref{fig:analysis:best} and Table \ref{fig:data} show that self-training annotations consist of  
shorter sentences and shallower trees than our dataset's average, i.e., mostly of easier sentences.

\begin{table}[t!]
\small
 \begin{tabular}{l R{1cm} R{1cm} R{1cm} R{1cm}}
\toprule
 & Av. Length & Av. Depth & Av. F1 & \#sents\\
 \midrule
Self-training & 7.0 & 3.3 & 82.2 & 1897\\
PTB gold & 20.9 & 10.6 & 100.0 & 39701\\
\bottomrule
\end{tabular}
\caption{Statistics of our best self-training annotations compared to PTB.}
\label{fig:data}

\end{table}

\begin{table}[t]
\small
\centering
 
\setlength{\tabcolsep}{2.0pt}
 \begin{tabular}{lrrrrr}
\toprule
Length & 0 - 10 & 10 - 20 & 20 - 30 & 30 - 40 & $>$ 40 \\
\midrule
Ex. & 115 & 573 & 613 & 295 & 94 \\
\% Ex. improved & 20\% & 36.8\% & 49.7\% & 52.8\% & 55.3\% \\
 \bottomrule
\end{tabular}
\caption{Percentage of development examples improved by SS-PRPN in comparison to PRPN, listed by sentence length.}
\label{fig:distribution}

\end{table}

\begin{figure}[]
\centering
\resizebox{0.8\columnwidth}{!}{%
\resizebox{6.5cm}{!}{
\begin{tikzpicture}
\begin{axis}[
    ylabel={Binary F1},
    xmin=50, xmax=250,
    ymin=40, ymax=65,
    xtick={50,100,150,200,250},
    ytick={40,45,50,55,60,65},
    legend pos=south east,
    ymajorgrids=true,
    grid style=dashed,
]

\addplot[
    color=blue,
    mark=square*,
    only marks
    ]
    coordinates {
    
  (50, 50.4)(100, 56)(150, 58.1)(200, 59.7)(250,61.8)
    };
    \addlegendentry{SP}
\addplot[
    color=gray,
    ]
    coordinates {
  (50, 46.3)(100, 46.3)(150, 46.3)(200, 46.3)(250, 46.3)
    };
    \addlegendentry{PRPN}
\addplot[
    color=black,
    ]
    coordinates {
  (50, 52.82)(100, 52.82)(150, 52.82)(200,52.82)(250,52.82)
    };
    \addlegendentry{C-PCFG}
\addplot[
    color=brown,
    ]
    coordinates {
  (50, 49.5)(100, 49.5)(150, 49.5)(200,49.5)(250, 49.5)
    };
    \addlegendentry{URNNG}
    
\addplot[
    color=green,
    mark=square*,
    only marks
    ]
    coordinates {
  (50, 45.4)(100, 47.9)(150, 56.4)(200,55.2)(250,52.5)
    };
    \addlegendentry{RNNG}
\addplot[
    color=red,
    mark=triangle*,
    only marks
    ]
    coordinates {
  (50,55.9)(100,57.3)(150, 58.7)(200,59.23)(250,59.4)
    };
    \addlegendentry{SS-PRPN}

\end{axis}

\end{tikzpicture}
}
}
\resizebox{0.8\columnwidth}{!}{%
\resizebox{6.5cm}{!}{
\begin{tikzpicture}
\begin{axis}[
    ylabel={Binary F1},
    xmin=50, xmax=250,
    ymin=35, ymax=55,
    xtick={50,100,150,200,250},
    ytick={35,40, 45,50, 55},
    legend pos=south east,
    ymajorgrids=true,
    grid style=dashed,
]

\addplot[
    color=blue,
    mark=square*,
    only marks
    ]
    coordinates {
  (50, 39.5)(100, 42.8)(150, 44.8)(200,46.8)(250,48.5)
    };
    \addlegendentry{SP} 
\addplot[
    color=red,
    mark=triangle*,
    only marks
    ]
    coordinates {
  (50, 46.6)(100, 48.2)(150, 49.5)(200, 49.8)(250,49.9)
    };
    \addlegendentry{SS-PRPN}

\end{axis}

\end{tikzpicture}
}
}
\caption{Low-resource parsing on the PTB. The first and second plots show unlabeled and labeled F1 respectively, plotted against the training data size.}
\label{fig:english}
\end{figure}
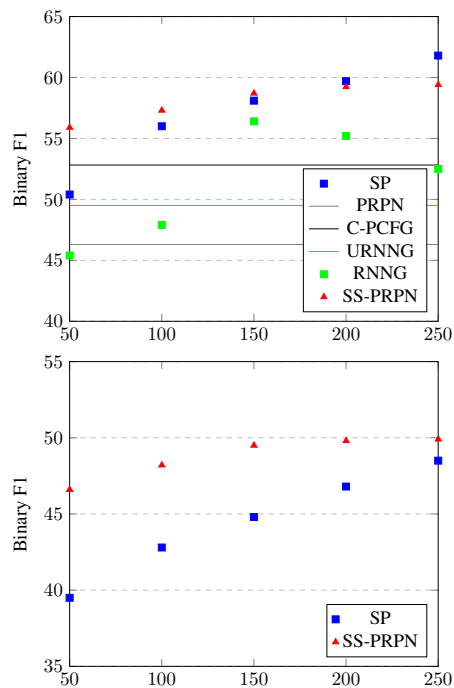
Our final hypothesis is that self-training helps mostly for longer sentences, since models often agree on shorter ones anyways and, trivially, longer sentences leave more room for error. 
Table \ref{fig:distribution} shows the development set performance and the number of examples for varying sentence lengths. 
As expected, self-training yields the greatest gains for longer sentences.


\paragraph{Low-Resource Parsing Performance}
We further investigate how SS-PRPN performs when limited gold parses are available in addition to unlabeled data. 
To predict constituency labels, we add and train an additional linear output layer
 after the first convolutional layer.
We find that, on the development set, converting $D_g$ into parse trees works better for low-resource parsing than $D_l$.
As supervised baselines, we employ \citet{dyer2016recurrent}'s  recurrent neural network grammar (RNNG) and a supervised parser (SP) based on syntactic distances \citep{shen2018straight}.
Figure \ref{fig:english} shows results for 50 to 250 annotated examples. The upper part shows the \textit{unlabeled} parsing performance in comparison to the UP baselines. We outperform all baselines for 50 to 150 examples, while SP performs slightly better with more annotations.
When looking at \textit{labeled} F1 in the lower part of Figure \ref{fig:english}, SS-PRPN clearly outperforms SP, which indicates that unlabeled data can be leveraged in the low-resource setting. 

\section{Conclusion}
We introduce a semi-supervised neural architecture, SS-PRPN, which is  capable of UP via self-training. Our self-trained models strongly outperform comparable baselines, and advance the state of the art on  PTB by $1.6\%$ F1. Analyses show that our approach yields most gains for longer sentences. Our architecture can also leverage limited amounts of parsing supervision 
when available. We conclude that it is beneficial to develop better UP models for semi-supervised settings.

\section{Acknowledgements}

This work has benefited from support of Samsung Research through the project \textit{Improving Deep Learning using Latent Structure} and the donation of Titan V GPU by NVIDIA Corporation. 


\bibliography{acl2020}
\bibliographystyle{acl_natbib}
\end{document}